\pdfoutput=1

\documentclass[11pt]{article}

\usepackage[final]{acl}

\usepackage{times}
\usepackage{latexsym}

\usepackage[T1]{fontenc}

\usepackage[utf8]{inputenc}

\usepackage{microtype}

\usepackage{inconsolata}

\usepackage{graphicx}
\usepackage{multirow}
\usepackage{makecell}
\usepackage{colortbl}
\usepackage{color}
\title{LightVLM: Acceleraing Large Multimodal Models with Pyramid Token Merging and KV Cache Compression}

\author{
 \textbf{Lianyu Hu\textsuperscript{1}},
 \textbf{Fanhua Shang\textsuperscript{1}},
 \textbf{Wei Feng\textsuperscript{1}},
 \textbf{Liang Wan\textsuperscript{1}},
\\
 \textsuperscript{1} College of Intelligence and Computing, Tianjin University, Tianjin, 300350, China,
\\
 \small{
   \textbf{Correspondence:} \href{Liang Wan}{Liang Wan:lwan@tju.edu.cn}
 }
}

\begin{document}
\maketitle
\begin{abstract}
In this paper, we introduce LightVLM, a simple but effective method that can be seamlessly deployed upon existing Vision-Language Models (VLMs) to greatly accelerate the inference process in a training-free manner. We divide the inference procedure of VLMs into two stages, i.e., encoding and decoding, and propose to simultaneously accelerate VLMs in both stages to largely improve model efficiency. During encoding, we propose pyramid token merging to reduce tokens of different LLM layers in a hierarchical manner by finally only keeping a few dominant tokens to achieve high efficiency. During decoding, aimed at reducing the high latency of outputting long sequences, we propose KV Cache compression to remove unnecessary caches to increase the network throughput. Experimental results show that LightVLM successfully retains 100\% performance when only preserving 35\% image tokens, and maintains around 98\% performance when keeping only 3\% image tokens. LightVLM could 2.02$\times$ the network throughput and reduce the prefilling time by 3.65$\times$. LightVLM also makes large VLMs faster again by enabling a heavy model (e.g., InternVL2.5 26B) to infer faster than significantly smaller models (e.g., InternVL2.5 8B), hopefully facilitating the real-world deployment. When generating long text sequences (e.g., 4096 tokens), LightVLM could reduce the inference time by 3.21$\times$, largely outperforming existing methods.  
\end{abstract}

\section{Introduction}
Recently, Vision-Language Models (VLMs)~\cite{li2024onevision,wang2024qwen2,bai2025qwen2,chen2024expanding,wang2024internvideo2} have gained rapid progress benefitted from the reasoning power of Large Language Models (LLMs)~\cite{dubey2024llama,yang2024qwen2,jiang2024mixtral}, which have demonstrated tremendous generalizability across a wide stream of vision-language tasks including image captioning~\cite{bai2025qwen2,zhang2025intra}, image question answering~\cite{chen2024expanding,li2024llavamed,li2023blip} and visual-text translation~\cite{hu2023continuous,lai2024lisa,hu2024scalable}. In these impressive models, images are usually first encoded into token sequences and concatenated with system prompts and user instructions, then fed into a LLM to generate desired textual outputs. The procedure of current VLMs can be divided into two stages: (1) encoding, VLMs process input system prompts, image tokens and user instructions as a whole; (2) decoding, based on encoded features and previous output tokens, recursively predicts the next token. Through cross-modal feature alignment and instruction tuning~\cite{liu2024visual,liu2024llavanext}, these VLMs could successfully adapt the perception capabilities from LLMs for vision reasoning, largely expanding the boundaries of existing vision models. 

\begin{figure}[t]
    \centering
    \includegraphics[width=\linewidth]{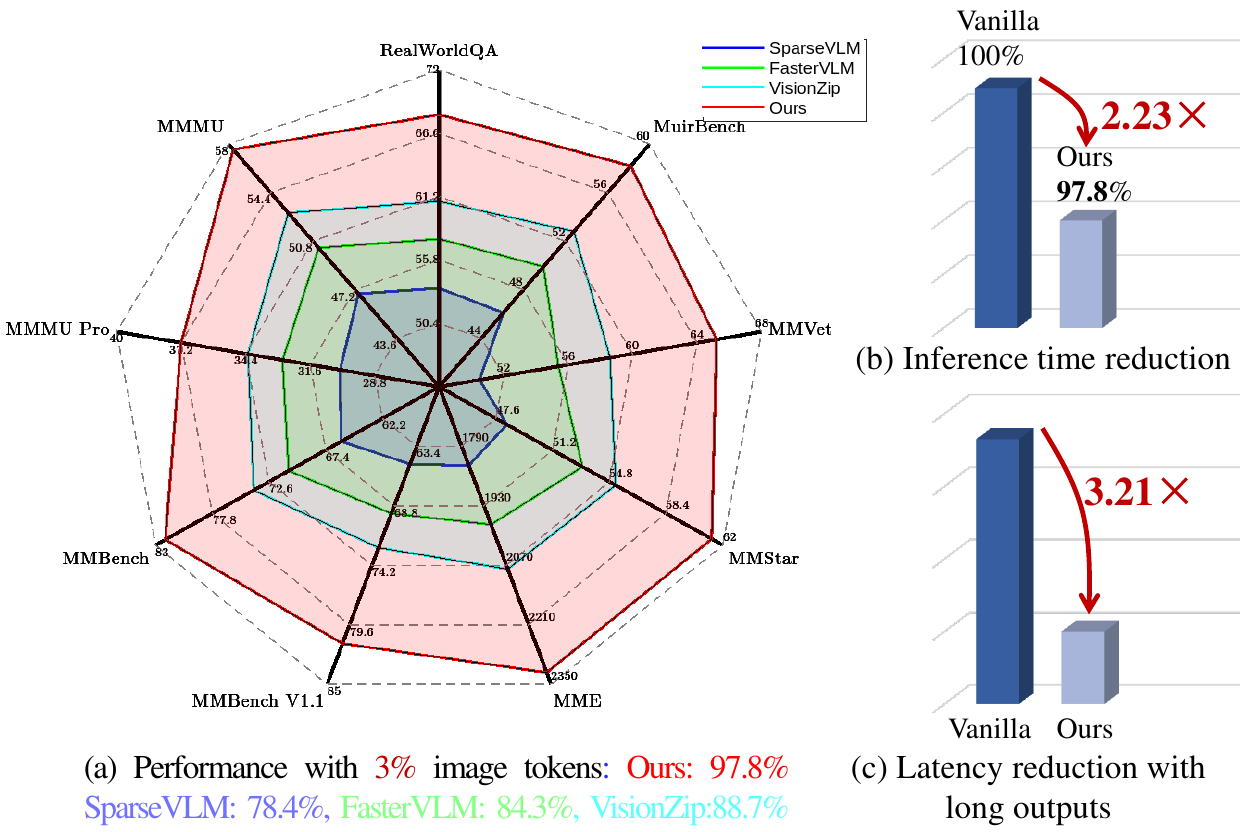} 
    \caption{(a) When retaining only 3\% image tokens, LightVLM largely outperforms existing methods over 9 benchmarks. (b) LightVLM could notably reduce inference time by 2.02$\times$ while maintaining 97.8\% performance. (c) LightVLM decreases response time by 3.21$\times$ when outputting long text sequences.}
    \label{fig1}
    \vspace{-15px}
  \end{figure}

However, the impressive performance of existing VLMs largely relies on the abundance of input image tokens during encoding. For example, the state-of-the-art VLM QWen2.5-VL~\cite{bai2025qwen2} usually transforms an image into thousands of tokens and accepts up to 16384 image tokens for processing. While the textual input just consists of a few dozens or hundreds of tokens, the vision side constantly dominates over 95\% or even 99\% input tokens. This constitutes a heavy burden in the processing procedure of VLMs, limiting their practical applications in real-world scenarios such as edge devices, autonomous driving and VR/AR computing~\cite{yu2024edge, jin2024efficient, yang2023lidar,gao25,gao2025knowledge,GAO2024106513,hu2024deep}. As previous works have revealed that there exists considerable redundancy in image sequences, we want to answer the question of how to better accelerate VLMs by squeezing image information to offer high efficiency in the encoding process.

When generating long sequences during decoding, KV Cache~\cite{bai2025qwen2,dubey2024llama,jiang2024mixtral} is widely utilized to accelerate models by caching the calculated key and value tensors of previous tokens in advance when predicting the next token, avoiding repeated computations. While it notably improves decoding speeds, the cached memory drastically increases as output sequences expand and becomes a heavy burden. For example, as output sequences grow from 128 to 8192 in VLMs, it brings extra 31.2G memory usage, which is unaffordable in many real-world scenarios. Besides, though already calculated key and value tensors are cached, they still need to compute attention scores with the current token during decoding, incurring expensive computations. We actively explore how to lower the high latency of VLMs when decoding long text sequences.

To handle the above challenges, we propose LightVLM, a simple but effective framework that can be seamlessly deployed upon VLMs for acceleration in a training-free manner. Our method accelerates both the encoding and decoding stages of VLMs. During encoding, we propose pyramid token merging to gradually reduce image tokens in a hierarchical manner, finally only preserving a few dominant image tokens (e.g., 3\%) without severely influencing the model performance. During decoding, we propose KV Cache compression to reduce required cache lengths by predicting and selecting the important image tokens before generation, decreasing the high latency brought by long output sequences. Results on 10+ benchmarks show that LightVLM could retain 100\% performance when keeping 35\% image tokens, and maintain 98\% performance when only keeping 3\% image tokens. LightVLM could decrease the inference time by 2.02$\times$, and reduce the latency by 3.21$\times$ when decoding long sequences. It successfully makes large VLMs faster again by enabling a larger model (e.g., 26B) to perform better and faster than smaller models (e.g., 8B). Comparison with existing efficient methods across different VLMs of various scales including 7B, 26B and 38B verify the effectiveness and flexibility of LightVLM.

\section{Related Work}
\subsection{Large Vision-Language Models} 
Our work is closely related to the surge of LVLMs. Traditional methods~\cite{ramachandram2017deep,xu2023multimodal,ochoa2017multimodal} usually collected large vision-language datasets and learned joint representation between vision and language from scratch to handle different tasks. These methods usually worked well in in-domain data but performed inferior in out-domain data that require common sense and world knowledge. 

Later, powered by the abundance of high-text data, LVLMs~\cite{li2024onevision,wang2024qwen2,alayrac2022flamingo,li2023blip,gao2024} shows impressive performance across a wide range of image understanding~\cite{yue2024mmmu,liu2024mmbench,masry2022chartqa,} and reasoning tasks~\cite{li2024seed,chen2024we}. These methods usually first transform the input images as patches, and then feed them into a vision-transformer-based image encoder. The extracted features are sent into a projector for dimension projection, whose outputs are further concatenated with the system prompts and user instructions to serve as inputs for the language model to generate textual outputs. As the attention mechanism with computational complexity $\mathcal{O}(n^2)$ is both used in the image encoder and language model in LVLMs, they have to consume high computational resources and own high inference latency when faced with long input sequences. 

\subsection{Token reduction}
Token reduction has been widely explored in both computer vision~\cite{meng2022adavit,chen2023sparsevit,pan2022less,ryoo2021tokenlearner,rao2021dynamicvit} and natural language processing (NLP)~\cite{goyal2020power,kim2020length,lassance2021study}. However, these methods usually require training, while our method can be done in a training-free manner. In multimodal learning, a series of methods tried to prune the tokens of intermediate layers for model acceleration. FastV~\cite{chen2024image} proposes to select the TopK-activated tokens within the language model to accelerate the forward pass. HiRED~\cite{arif2024hired} presents a dynamic high-resolution early dropping strategy for allocating computing resources between partitioned sub-images and the main image. Zhang et al.~\cite{zhang2024token} propose to identify redundant tokens by assessing the pattern repetitiveness of correlations between patches and then conducting pruning. LLaVA-PruMerge~\cite{shang2024llava} leverages the attention scores between the [CLS] token and other tokens in the image encoder to guide token dropping before processing with the language model. SparseVLM~\cite{zhang2025sparsevlm} proposes a rank-based strategy to adaptively determine the sparsification ratio for each layer to reduce image tokens. FasterVLM~\cite{zhang2024cls} utilizes attentions between the [CLS] token and image tokens from the visual encoder to decrease imge tokens in the image encoder. VisionZip~\cite{yang2025visionzip} introduces a token pruning and mergeing strategy to compress image tokens after the image encoder according to attention maps from visual encoders. Compared to these methods, our method show clear advantages over accuracy across different benchmarks. Our method demonstrates more superior performance with less image tokens.
\subsection{KV Cache Compression}
The exploration of efficient modeling has been broadly explored in neural language modeling (NLP) fields~\cite{wan2024efficient,wan2025d2o}. For example, SnapKV~\cite{li2024snapkv} proposes to compress the KV Cache to a constant length across different layers for compression. PyramidKV~\cite{zefan2024pyramidkv} proposes to dynamically allocate the cache resources across different layers with a predefined scehdule. MEDA~\cite{wan2025meda} utilizes cross-modal attention entropy to determine the KV cache size at each MLLMs layer with a KV pair selection scheme for token merging. CAKE~\cite{qin2025cake} first assesses layer-specific preferences in both spatial and temporal dimensions, and then employs a new eviction indicator to consider the shifting importance of tokens. Up to now, few works have explored how to accelerate VLMs in the decoding stage by compressing KV Cache, and we hope we could make a meaningful attempt. 

\section{Method}
\subsection{Our Observations}
\label{sec2.1}
\begin{figure*}[t]
    \centering
    \includegraphics[width=\linewidth]{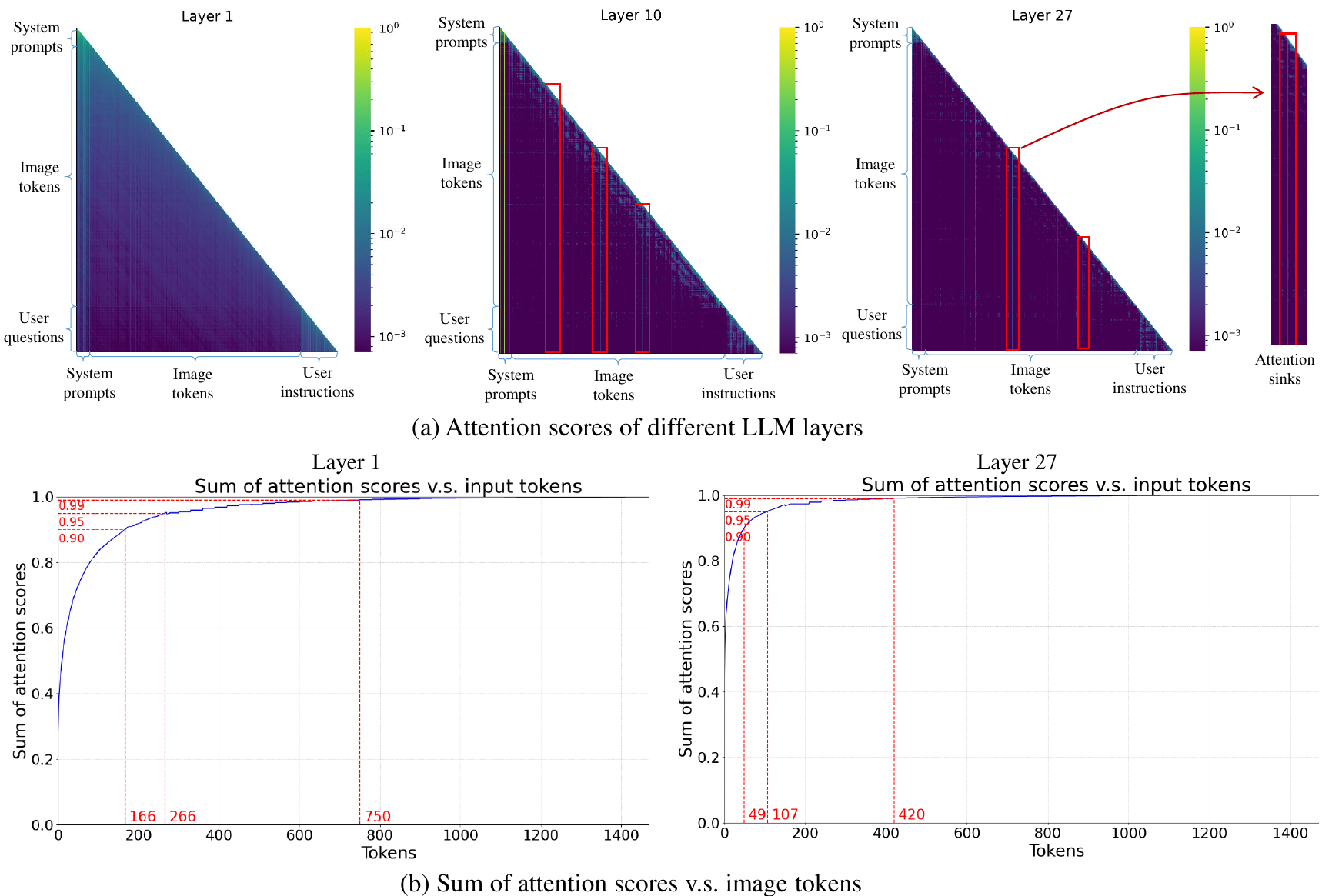} 
    \caption{(a) The attention score distribution for input tokens including system prompts, image tokens and user instructions when encoding each token in different LLM layers. (b) The sum of attention scores corresponds to the number of image tokens in different LLM layers.}
    \label{fig2}
\end{figure*}

Previous methods~\cite{chen2024image,zhang2025sparsevlm} have revealed that there exists considerable feature redundancy in images. In this paper, we conduct a deeper study of feature redundancy in VLMs, which directly motivates the invention of our method. We plot the pilot results in Fig.~\ref{fig2}(a) and Fig.~\ref{fig2}(b), respectively.

We first investigate the correlations between image inputs and textual inputs. Fig.~\ref{fig2}(a) draws the attention score distribution over input tokens including system prompts, image tokens and user instructions when encoding each token in VLMs. We have the following findings. (1) We find textual inputs (e.g., system prompts and user instructions) receive much more attention than image tokens across all LLM layers (Layer 1 - Layer 27), which reflects that the information is much sparser in images than input texts. (2) For the input image itself, the attention scores received by image tokens are generally uniform in shallow layers (Layer 1 in Fig.~\ref{fig2}(a)), which gradually converge to some dominant image tokens as layers deepen (Layer 10 and 27 in Fig.~\ref{fig2}(a)). It seems VLMs first try to encode beneficial information from each input image token in earlier layers, and finally aggregate and consolidate key information from several key image tokens in deeper layers. (3) We observe there exist `attention sinks' in deeper layers (red boxes for Layer 10 and 27 in Fig.~\ref{fig2}(a)). Some tokens are activated all the time when processing each token. We speculate that VLMs may condense beneficial information into these tokens, which undertake the role of providing guidance or expert knowledge for other tokens.

We then explore the internal relationships within image tokens. Fig.~\ref{fig2}(b) shows the sum of attention scores corresponding to the number of image tokens. We notice that few image tokens dominate most attention scores. In Layer 1, 11\% image tokens (188/1476) dominate 95\% scores, and 18\% image tokens (266/1476) own 99\% scores. The trend is more biased in deep layers. In Layer 27, just 49 image tokens (3\%) take up 90\% attention scores, while \textless 10\% image tokens (107/1476) consume 95\% attention scores. Within the image itself, the information is unevenly distributed and most image tokens are neglected in deep layers.

From above observations, we draw two major conclusions: (1) Image tokens receive much less attention than textual inputs (e.g., system prompt, user instructions) in VLMs, which further decrease as layers deepen. (2) Within image tokens, few tokens dominate the attention scores and the distribution is more sinked as layers go deeper. This motivates us to explore how to better reduce image feature redundancy to accelerate models.

\subsection{LightVLM}
\begin{figure*}[t]
    \centering
    \includegraphics[width=0.85\linewidth]{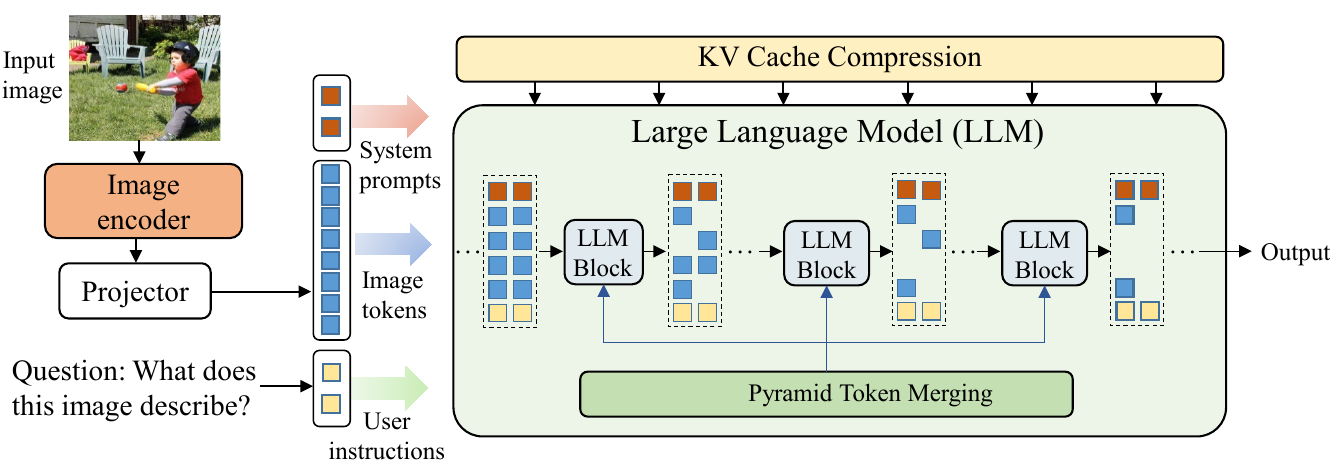} 
    \caption{The input image is first transformed into image token sequences by an image encoder and a projector, which are then concatenated with system prompts and user instructions as the inputs for LLM to perform reasoning. We perform pyramid token merging in certain layers of LLM to accelerate models, and conduct KV Cache compression for each LLM layer to accelerate the generation stage.}
    \label{fig3}
    \vspace{-8px}
\end{figure*}

We propose LightVLM, a lightweight framework that can be seamlessly deployed upon VLMs in a training-free manner to perform acceleration, whose framework is shown in Fig.~\ref{fig3}. In popular VLMs, given an input image, it's first transformed into image token sequences by an image encoder and a projector, which are then concatenated with system prompts and user instructions as the inputs for LLM to perform reasoning. The procedure of current VLMs can be divided into two stages: (1) encoding, VLMs process input system prompts, image tokens and user instructions as a whole; (2) decoding, based on encoded features and previous output tokens, recursively predicts the next token. We here accelerate VLMs from two perspectives: (1) during the encoding stage, towards the heavy computations brought by high image feature redundancy, we propose pyramid token merging to gradually remove unnecessary image tokens across layers, finally only keeping a few image tokens. (2) during the decoding stage, we propose KV Cache compression to reduce network latency and decrease massive memory usage brought by long cache sequences. As shown in Fig.~\ref{fig3}, we apply pyramid token merging strategy in some specific layers LLM $K$, and activate KV Cache compression for each LLM layer after some earlier layers.

\subsubsection{Pyramid token merging}
According to our observations from Fig.~\ref{fig2}, we maintain textual input tokens across all layers and just perform token merging for image tokens to reduce feature redundancy. As in earlier layers (e.g., Layer 1 in Fig.~\ref{fig2}(a)) VLMs assign uniform attention scores to all image tokens, we keep all image tokens with no reduction in initial LLM layers, As layers deepen, we propose to reduce image tokens in a pyramid manner to only keep necessary information for certain LLM layers $K$ and discard others for high efficiency.


To preserve the informative tokens that are most important to the model, we first need to clarify how to measure the importance of each token. Based on observations from Fig.~\ref{fig2}, we find that the attention scores from the LLM serve as a good indicator for the significance of image tokens, as preserving image tokens with the highest scores could well maintain the model performance. We thus adopt the attention scores as our metrics.

With token importance defined, we need a fast way to perform token merging to reduce image tokens per layer. Before designing specific token merging algorithms, there are several key problems in practical scenarios that deserve careful considerations. First, while the dot-producted attention is most well-known and adopted as the default configuration in standard Transformer architectures in VLMs, efficient attention approaches, especially Flash-Attention, are quickly developing and more and more adopted in recent VLMs. They even become the standard configurations in state-of-the-art VLMs. However, different from the vanilla attention that computes attention scores $A\in \mathcal{R}^{N\times N}$ with $N$ representing the sequence length, Flash-Attention only returns cumsum attention scores\footnote{We can pass the parameter $return\_attn\_probs$ to $flash\_attn\_varlen\_func$ function to enable returning cumsum attention weights $A\in \mathcal{R}^{N}$.}  $A\in \mathcal{R}^{N}$ due to split computation procedure. Here, we omit the batch size and use the averaged attention scores across diferent heads for clarify. The designed token merging algorithm needs to be compatible with efficient attention approaches as much as possible to avoid incurring additional massive computations. Second, we perform token merging operations multiple times within the network on potentially thousands of tokens. The runtime of this algorithm has to be absolutely negligible. Thus, we want to avoid anything that cannot be parallelized or is iteratively computed. While some clustering approaches are preferred by recent works~\cite{zhang2025sparsevlm}, they inevitably introduce additional running latency.


Our token merging algorithm is as follows. First, we divide the image tokens into two subsets $P_{\rm unmerged}\in \mathcal{R}^{(N - (r+1))\times C}$ with length of $N - (r+1)$ and $P_{\rm merged}\in \mathcal{R}^{(r+1)\times C}$ with a length of $r+1$ based on the attention scores $A\in \mathcal{R}^{N}$ returned by Flash-Attention, with $r$ denoting the token reduction number and $C$ representing the channels. We select $P_{\rm unmerged}$ as tokens with the highest attention scores, and apply token merging for $P_{\rm merged}$ only. We first reorder the tokens in $P_{\rm merged}$ by their importance, with tokens of the least importance at the end of the queue. Next, considering tokens in the end of the queue are the least important for the model, we combine the last two tokens into a new robust representation, via merging their values with equal weights. We repeatedly conduct this procedure until meeting the required token budget. As the merging process is deterministic after obtaining the attention scores for $P_{\rm merged}$, this process can be simplified to a simple multiplication between tokens in $P_{\rm merged}$ with a pre-calculated weight list $W_v\in {\mathcal{R}^{r+1}} = \{r+1, r, \dots, 1\}$. Thus, we can finish the merging process with a simple multiplication to obtain $P_{\rm new}\in \mathcal{R}^{1\times C} = W_v \times P_{\rm merged}$ once, with only $\mathcal{O}(1)$ computational complexity. After the merging process, $P_{\rm new}$ is concatenated with $P_{\rm unmerged}$ to form the compressed token sequences $P_{\rm compressed} \in \mathcal{R}^{(N-r)\times C}$ = [$P_{\rm unmerged}$, $P_{\rm new}$], with length of $N - r$. The token merging procedure doesn't require the full attention matrix $A\in \mathcal{R}^{N\times N}$ and can be completely finished with only the cumsum attention scores $A\in \mathcal{R}^{ N}$ returned by Flash-Attention.

Overall, we expect to perform pyramid token merging in early, middle and later LLM layers to keep low-level, mid-level and high-level visual information to maintain necessary semantics with high efficiency as much as possible.

\begin{table*}[t]
    \small	
    \setlength\tabcolsep{1pt}
    \centering
    \caption{Results on image benchmarks with QWen2.5-VL 7B over different kept ratios. }
    \vspace{-5px}
    \label{tab2}
    \begin{tabular}{c|ccccccccc|c}
    \hline Method & \makecell[c]{MMMU \\(Val)} & \makecell[c]{MMMU Pro\\Overall} & \makecell[c]{MMBench \\(EN)} & \makecell[c]{MMBench V1.1\\(EN)}&  \makecell[c]{MME\\(sum)} & \makecell[c]{ MMStar \\(Test)} & \makecell[c]{MMVet\\(Test)} & \makecell[c]{MuirBench} & RealWorldQA & Avg \\
    \hline
    \multirow{2}{*}{\makecell{Vanilla \\(Upper bound)}} & 58.6 & 38.3 & 83.5 & 82.6 & 2347 & 63.9 & 67.1 & 59.6 & 68.5 & 65.3\\
    & 100.0\% & 100.0\% & 100.0\% & 100.0\% & 100.0\% & 100.0\% & 100.0\% & 100.0\%  & 100.0\%  & 100.0\% \\
    \hline
    \rowcolor{gray!20}
    & \multicolumn{9}{c|}{Keep \textcolor{red}{35\%} image tokens } & \\
    \multirow{2}{*}{SparseVLM} & 56.4& 36.8& 80.0& 79.0& 2263& 60.6& 63.5& 57.0& 65.6 & 62.5\\
    & 96.3\% & 96.2\% & 95.8\% & 95.7\% & 96.4\% & 94.8\% & 94.7\% & 95.6\% & 95.7\% & 95.7\%\\
    \hline
    \multirow{2}{*}{FasterVLM} & 57.7& 37.6& 81.7& 80.9& 2298& 62.1& 64.9& 58.6& 67.5 & 64.0\\
    & 98.4\% & 98.2\% & 97.8\% & 98.0\% & 97.9\% & 97.2\% & 96.7\% & 98.4\% & 98.5\% & 98.0\%\\
    \hline
    \multirow{2}{*}{PyramidDrop} &58.1 & 38.0 & 82.3 & 81.6 & 2302& 62.1 & 65.0 & 58.5 & 67.4 & 64.1 \\
    & 99.2\% & 99.3\% & 98.5\% & 98.8\% & 98.1\% & 97.2\% & 96.8\% & 98.1\% & 98.4\% & 98.2\% \\
    \hline
    \multirow{2}{*}{VisionZip} & 57.8 & 37.8 & 80.0 & 81.2 & 2305 & 62.7 & 66.1 & 58.6 & 67.3 & 64.3\\
    & 98.7\% & 98.6\% & 95.8\% & 98.3\% & 98.2\% & 98.1\% & 98.5\% & 98.4\% & 98.3\% & 98.5\% \\
    \hline
    \multirow{2}{*}{LightVLM} & 58.7 & 38.5 & 83.6 & 82.5 & 2352 & 63.8 & 67.1 & 59.6 & 68.6 & \textbf{65.3}\\
    & 100.2\%& 100.5\%& 100.1\%& 99.9\%& 100.2\%& 99.8\%& 100.0\%& 100.0\%& 100.1\% & \textbf{100.0\%}\\
    \hline
    \rowcolor{gray!20}
    & \multicolumn{9}{c|}{Keep \textcolor{red}{15\%} image tokens } & \\
    \multirow{2}{*}{SparseVLM} & 52.3& 34.2& 73.8& 73.3& 2065& 55.4& 58.2& 52.3& 60.3 & 57.6\\
    &89.3\% & 89.2\% & 88.4\% & 88.7\% & 88.0\% & 86.7\% & 86.8\% & 87.8\% & 88.1\% &  88.2\% \\
    \hline
    \multirow{2}{*}{PyramidDrop} & 54.9 & 35.7 & 76.8 & 76.5 & 2140 & 57.1 & 60.3 & 53.3 & 61.9 & 59.5\\
    &  93.7\% & 93.1\% & 92.0\% & 92.6\% & 91.2\% & 89.3\% & 89.9\% & 89.5\% & 90.3\% & 91.2\%\\
    \hline
    \multirow{2}{*}{FasterVLM} & 55.8& 36.5& 79.1& 78.1& 2211& 59.5& 62.3& 57.0& 65.3 &61.7\\
    & 95.2\% & 95.3\% & 94.7\% & 94.6\% & 94.2\% & 93.1\% & 92.8\% & 95.6\% & 95.4\% & 94.5\%\\
    \hline
    \multirow{2}{*}{VisionZip} & 56.0& 36.5& 79.5& 78.6& 2232& 60.5& 63.5& 56.6& 65.1 & 62.1\\
    & 95.6\% & 95.4\% & 95.2\% & 95.2\% & 95.1\% & 94.7\% & 94.6\% & 95.0\% & 95.1\%  & 95.1\% \\
    \hline
    \multirow{2}{*}{LightVLM} & 58.4 & 37.8 & 83.2 & 82.4 & 2342 & 62.1 & 65.5 & 59.1 & 68.4 & \textbf{64.6}\\
    & 99.7\%& 98.7\%& 99.6\%& 99.8\%& 99.8\%& 97.2\%& 97.6\%& 99.2\%& 99.9\% & \textbf{98.9\%}\\
    \hline
    \rowcolor{gray!20}
    & \multicolumn{9}{c|}{Keep \textcolor{red}{3\%} image tokens } & \\
    \multirow{2}{*}{PyramidDrop} & 46.7 & 30.2 & 63.5 & 65.6 & 1783 & 46.8 & 49.9 & 42.9 & 49.2 & 49.4 \\
    & 79.6\% & 78.9\% &76.0\% & 79.3\% & 76.0\% & 73.2\% & 74.5\% & 72.1\% & 71.8\% & 75.7\%\\
    \hline
    \multirow{2}{*}{SparseVLM} & 46.9& 30.3& 66.0& 65.0& 1835& 48.3& 50.5& 46.1& 53.4 & 51.2\\
    & 80.1\% & 79.2\% & 79.1\% & 78.7\% & 78.2\% & 75.6\% & 75.2\% & 77.4\% & 78.0\% &  78.4\%\\
    \hline
    \multirow{2}{*}{FasterVLM} & 50.3& 32.8& 70.8& 69.5& 1974& 53.1& 55.4& 49.9& 57.6 & 55.0 \\
    & 85.8\% & 85.6\% & 84.8\% & 84.2\% & 84.1\% & 83.1\% & 82.5\% & 83.8\% & 84.1\% & 84.3\%\\
    \hline
    \multirow{2}{*}{VisionZip} & 52.9& 34.3& 74.0& 72.6& 2079& 55.2& 58.6& 52.8& 60.8 & 57.9 \\
    & 90.2\% & 89.5\% & 88.6\% & 87.9\% & 88.6\% & 86.4\% & 87.3\% & 88.6\% & 88.7\% & 88.7\%\\
    \hline
    \multirow{2}{*}{LightVLM} & 57.6 & 37.2 & 82.1 & 81.3 & 2322 & 61.3 & 65.2 & 58.2 & 68.2 & \textbf{63.9} \\
    & 98.3\%& 97.1\%& 98.3\%& 98.4\%& 98.9\%& 95.9\%& 97.2\%& 97.7\%& 99.6\% & \textbf{97.8\%}\\
    \hline
    \end{tabular}
    \vspace{-7px}
\end{table*}

\subsubsection{KV Cache compression}
KV Cache~\cite{bai2025qwen2,dubey2024llama,jiang2024mixtral} is widely adopted in the decoding stage of LLMs and VLMs to store calculated key and values of previous output tokens to avoid repeated computing when predicting the next token. While KV Cache effectively reduces required computations, the drastically increased memory usage caused by cached tensors when decoding sequences poses a significant challenge for memory-constrained scenarios. Besides, the current token still needs to calculate attention scores with all cached keys and values, incurring expensive computations. To lower the network latency by tackling these limitations, we propose to compress the KV Cache in VLMs.

Fig.~\ref{fig2}(b) reveals that few image tokens (e.g., \textless 10\%) dominate (e.g., 95\%) the attention scores in VLMs, and the attention distribution is even steeper in deeper layers. For each LLM layer (except shallow layers), we leverage the attention scores to only keep a few image tokens with the highest attention score for caching. The other image tokens are discarded and not used in the subsequent generation stage, thereby saving memory usage and boosting network throughput. Specifically, based on either the full attention matrix $A$ returned by the vanilla attention or the cumsum attention returned by Flash-Attention, we first calculate the attention scores $A$ corresponding to each token, and then sort it in a descending manner with tokens of the highest importance at the front to obtain $A_{\rm des}$. Next, we keep image tokens with accumulated attention score above the predefined threshold $\beta$, and directly discard others in the following LLM layers. We prune image tokens in the KV Cache with the same attention threshold for different layers. However, as reflected by Fig.~\ref{fig2}, since VLMs tend to assign uniform/spiky attention to image tokens in shallow/deep LLM layers, this strategy tends to keep more tokens in shallow LLM layers while preserving fewer in deeper layers. It's worth noting that our KV Cache compression is conducted independently per head in each layer, offering high flexibility for VLM computing. 

\section{Experimental Results}

\begin{table}[t]
    \fontsize{8pt}{0.7\baselineskip}\selectfont
    \setlength\tabcolsep{1pt}
    \centering
    \caption{Results on video benchmarks with QWen2.5-VL 7B over different kept ratios. EgoSchema is abbreviated as EgoS.}
    \label{tab3}
    \begin{tabular}{c|cccc|c}
    \hline Method & \makecell[c]{VideoMME \\ (w/wo sub))} & \makecell[c]{MVBench} & \makecell[c]{EgoS \\(Test)} & \makecell[c]{MLVU\\(M-Avg)} & Avg \\
    \hline
    \multirow{2}{*}{\makecell[c]{Vanilla \\ (Upper bound)}} & 71.6 / 65.1 & 69.6 & 65.0 & 70.2 & 68.3 \\
    & 100.0\%/100.0\%& 100.0\%& 100.0\%& 100.0\%& 100.0\% \\
    \hline
    \rowcolor{gray!20}
    & \multicolumn{4}{c|}{Keep \textcolor{red}{35\%} video tokens } & \\
    \multirow{2}{*}{SparseVLM} & 68.4 / 62.2& 66.5& 62.1& 67.0 & 65.5\\
    & 96.5\% / 96.4\% & 96.4\% & 95.7\% & 95.6\% & 95.9\%\\
    \hline
    \multirow{2}{*}{FasterVLM} & 69.8 / 63.5& 67.9& 63.4& 68.4 & 66.6 \\
    & 97.8\% / 97.6\% & 97.2\% & 97.6\% & 97.3\% & 97.5\%\\ 
    \hline
    \multirow{2}{*}{PyramidDrop}& 70.5/63.7 & 68.4 & 63.2 & 69.1 & 67.0 \\
    &98.5\%/97.9\% & 98.3\% & 97.2\% & 98.4\% & 98.0\%\\
    \hline
    \multirow{2}{*}{\makecell[c]{VisionZip}} & 70.5/64.0& 68.1& 63.7& 68.9 & 67.1\\
    & 98.5\%/98.3\% & 97.9\% & 98.0\% & 98.2\% & 98.3\%\\
    \hline
    \multirow{2}{*}{LightVLM} & 71.7 / 65.1 & 69.7 & 65.0 & 70.1 & \textbf{68.3}\\
    & 100.1\%/100.0\% & 100.1\% & 100.0\% & 99.8\% & \textbf{100.0\%} \\
    \hline
    \rowcolor{gray!20}
    & \multicolumn{4}{c|}{Keep \textcolor{red}{15\%} video tokens } & \\
    \multirow{2}{*}{SparseVLM} & 66.2 / 60.2& 64.4& 60.1& 64.9 & 63.2\\
    & 93.1\%/93.2\% & 92.8\% & 92.1\% & 92.2\% & 92.5\% \\
    \hline
    \multirow{2}{*}{PyramidDrop} & 67.4/60.9 & 65.6 & 60.1 & 65.0 & 63.7 \\
    & 94.2\%/93.6\% & 94.3\% & 92.4\% & 92.6\% & 93.2\% \\
    \hline
    \multirow{2}{*}{FasterVLM} & 67.3 / 61.2& 65.4& 61.1& 66.0 & 64.4\\
    & 94.8\%/94.6\% & 94.9\% & 94.1\% & 93.8\% & 94.3\% \\
    \hline
    \multirow{2}{*}{\makecell[c]{VisionZip}} &68.8/62.4& 66.4& 61.9& 66.8 & 65.2\\
    & 96.1\%/95.8\% & 95.4\% & 95.2\% & 95.1\% & 95.4\% \\
    \hline
    \multirow{2}{*}{LightVLM} & 71.3 / 65.0 & 69.4 & 64.1 & 69.6 & \textbf{67.8}\\
    & 99.6\%/99.8\% & 99.7\% & 98.9\% & 99.1\% & \textbf{99.3\%} \\
    \hline
    \rowcolor{gray!20}
    & \multicolumn{4}{c|}{Keep \textcolor{red}{3\%} video tokens } & \\
    \multirow{2}{*}{SparseVLM} & 58.1/52.6& 56.0& 52.0& 56.0 & 54.9\\
     & 81.2\%/80.8\% & 80.5\% & 79.8\% & 80.2\% &  80.4\% \\
    \hline
    \multirow{2}{*}{PyramidDrop} & 60.0/54.2 & 57.4 & 53.1 & 57.7 & 56.5 \\
    & 83.8\%/83.2\% & 82.4\% & 81.7\% & 82.2\% & 82.8\% \\
    \hline
    \multirow{2}{*}{FasterVLM} & 61.0 / 55.5& 59.3& 55.4& 59.8 & 52.5\\
    & 86.1\%/86.2\% &85.3\% &84.8\% &84.5\% &85.2\% \\
    \hline
    \multirow{2}{*}{\makecell[c]{VisionZip}} & 64.2 / 58.3& 62.4& 58.2& 62.9 & 61.1\\
    & 90.1\%/89.7\% & 89.3\% &89.1\% &88.7\% & 89.4\% \\
    \hline
    \multirow{2}{*}{LightVLM} & 70.5 / 64.1& 68.6& 64.0& 68.7 & \textbf{67.1} \\
    & 98.5\%/98.5\% & 98.5\% & 98.2\% & 99.7\% & \textbf{98.3\%} \\ 
    \hline
    \end{tabular}
    \vspace{-5px}
\end{table}

\subsection{Effectiveness on Image Understanding} 

We compare our LightVLM with PyramidDrop~\cite{xing2025pyramiddrop}, SparseVLM~\cite{zhang2025sparsevlm}, FasterVLM~\cite{zhang2024cls} and VisionZip~\cite{yang2025visionzip} over 9 image benchmarks~\cite{yue2024mmmu,yue2024mmmupro,liu2024mmbench,liu2024mmbench,xu2023lvlm,chen2024we,yu2023mm,wang2024muirbench,RealWorldQA} by deploying upon a state-of-the-art QWen2.5-VL 7B model~\cite{bai2025qwen2} in Tab.~\ref{tab2}. We compare them by maintaining comparably 35\%, 15\% or 3\% image tokens for fair comparisons.  LightVLM outperforms other efficient methods upon different kept ratios over all benchmarks. Especially, when retaining 35\% image tokens, LightVLM outperforms other methods by 1.0\%/1.5\% in absolute/relative performance. When the preserved ratio decreases to 15\% and 3\%,  the performance gaps between LightVLM and others drastically increase to 2.5\%/3.8\% and 6.0\%/9.1\% in absolute/relative performance. This reflects the superiority of LightVLM in retaining key vision information under extreme cases. An interesting phenomenon is that by reducing image tokens to 35\%, LightVLM leads to performance boost upon several benchmarks, which shows that considerably compressing image tokens can help VLMs better perform perception.

\begin{table}[t]
    \fontsize{8pt}{0.7\baselineskip}\selectfont
    \setlength\tabcolsep{1pt}
    \centering
    \caption{Flexibility of LightVLM upon various VLMs. We abbreviate RealWorldQA as RWQA here. }
    \vspace{-5px}
    \label{tab4}
    \begin{tabular}{c|cccc|c}
    \hline 
     & \makecell[c]{MMMU \\(Val)} & \makecell[c]{MMBench \\(EN Dev)} & \makecell[c]{MMStar\\ (Test)}&  \makecell[c]{RWQA \\ (Test)} & Avg \\
    \hline
    \hline
    \rowcolor{gray!20}
    LLaVA Onevision 7B & 48.8 & 80.8 & 61.7 & 66.3 & 64.4\\
    \hline
    SparseVLM & 46.8& 77.6& 59.2& 63.6 & 61.8\\
    FasterVLM & 47.6& 78.8 & 60.2& 64.6 & 62.8\\
    VisionZip & 48.0 & 79.2 & 60.4 & 65.1 & 63.2\\
    PyramidDrop & 48.2 & 79.8 & 61.0& 64.9 & 63.5\\
    LightVLM & \textbf{48.9} & \textbf{80.9}  & \textbf{61.6} & \textbf{66.3} & \textbf{64.4}\\
    \hline
    \hline
    \rowcolor{gray!20}
    InternVL2.5 8B & 56.0 & 84.6 & 62.8 & 70.1 & 68.4 \\
    \hline
    SparseVLM & 53.1& 81.3& 59.2& 65.6 & 64.8\\
    FasterVLM & 54.2& 82.5 & 60.2& 66.8 & 65.9\\
    PyramidDrop & 54.2 & 82.7 & 60.6 & 68.3 & 66.3\\
    VisionZip & 54.8 & 83.2 & 61.3 & 68.7 & 67.0\\
    LightVLM & \textbf{56.0} & \textbf{84.7} & \textbf{62.8} & \textbf{70.2} & \textbf{68.4}\\
    \hline
    \hline
    \rowcolor{gray!20}
    MiniCPM-V2.6 8B & 49.8 & 81.5 & 57.5 & 65.0 & 63.5\\
    \hline
    SparseVLM & 47.6& 77.8& 54.9& 62.1 & 60.6\\
    FasterVLM & 48.6& 79.5& 56.1& 63.4 & 61.9\\
    PyramidDrop & 48.0 & 80.3 & 55.4& 64.2 & 62.2\\
    VisionZip & 48.9 & 80.2 & 56.5 & 64.0 & 62.4\\
    LightVLM & \textbf{49.8} & \textbf{81.6} & \textbf{57.4} & \textbf{65.1} & \textbf{63.5}\\
    \hline
    \end{tabular}
\end{table}

\subsection{Effectiveness on Video Understanding}

We compare our LightVLM with PyramidDrop~\cite{xing2025pyramiddrop}, SparseVLM~\cite{zhang2025sparsevlm}, FasterVLM~\cite{zhang2024cls} and VisionZip~\cite{yang2025visionzip} by keeping 35\%, 15\% or 3\% image tokens upon the QWen2.5-VL 7B model~\cite{bai2025qwen2} over 4 video benchmarks including VideoMME~\cite{fu2024video}, MVBench~\cite{li2024mvbench}, EgoSchema~\cite{mangalam2023egoschema} and MLVU~\cite{zhou2024mlvu} in Tab.~\ref{tab3}. We observe that LightVLM consistently outperforms other methods across different benchmarks with various kept ratios, and demonstrates more superiority advantages when retaining fewer (15\% and 3\%) image tokens, validating its efficacy.

\subsection{Flexibility of LightVLM}

\textbf{Flexibility of LightVLM upon various VLMs.} We validate the flexibility of LightVLM by deploying it upon various VLMs including LLaVA Onevision 7B~\cite{li2024onevision}, InternVL2.5 8B~\cite{chen2024expanding} and MiniCPM-V2.6 8B~\cite{yu2024rlaif} in Tab.~\ref{tab4} on 4 image benchmarks. We observe that across different VLMs, LightVLM consistently maintains around 100\% performance and outperforms other efficient methods by a large margin, proving its flexibility and efficacy.

\textbf{Flexibility of LightVLM upon VLMs of various scales.} We validate the flexibility of LightVLM by deploying it upon VLMs of various scales including InternVL2.5 8B~\cite{chen2024expanding}, InternVL2.5 26Band InternVL2.5 38B in Tab.~\ref{tab5}. We observe that as a plug-and-play method, LightVLM could notably accelerate VLMs of different scales and outperform other efficient methods by a large margin across different benchmarks, demonstrating its strong generalizability.

\begin{table}[t]
    \small	
    \setlength\tabcolsep{1pt}
    \centering
    \caption{LightVLM with VLMs of various scales. We abbreviate RealWorldQA as RWQA here.}
    \vspace{-5px}
    \label{tab5}
    \begin{tabular}{c|cccc|c}
    \hline 
     & \makecell[c]{MMMU \\(Val)} & \makecell[c]{MMBench \\(EN Dev)} & \makecell[c]{MMStar\\ (Test)}&  \makecell[c]{RWQA \\ (Test)} & Avg \\
    \hline
    \hline
    \rowcolor{gray!20}
    InternVL2.5 8B & 56.0 & 84.6 & 62.8 & 70.1 & 68.4 \\
    \hline
    SparseVLM & 53.6& 81.0& 60.1& 67.1& 65.5\\
    FasterVLM & 54.7& 82.6& 61.3& 68.4& 66.8\\
    PyramidDrop & 54.2 & 82.7 & 60.6 & 68.3 & 66.3\\
    VisionZip & 55.1 & 83.1 & 61.7 & 69.0 & 67.2\\
    LightVLM & \textbf{56.0} & \textbf{84.7} & \textbf{62.8} & \textbf{70.2} & \textbf{68.4}\\
    \hline
    \hline
    \rowcolor{gray!20}
    InternVL2.5 26B & 60.0 & 85.4 & 66.5 & 74.5 & 71.6\\
    \hline
    SparseVLM & 57.4& 81.7& 63.6& 71.3& 68.5\\
    FasterVLM  & 58.6& 83.4& 64.9& 72.7& 69.9 \\
    VisionZip & 59.0 & 83.9 & 65.2 & 73.1 & 70.3 \\
    PyramidDrop & 59.8 & 83.7 & 65.6 & 73.0 & 70.5\\
    LightVLM & \textbf{60.1} & \textbf{85.4} & \textbf{66.6} & \textbf{74.6} & \textbf{71.6} \\
    \hline
    \hline
    \rowcolor{gray!20}
    InternVL2.5 38B & 63.9 & 86.5 & 67.9 & 73.5 & 73.0\\
    \hline
    SparseVLM & 61.0& 82.5& 64.8& 70.1& 69.6\\
    FasterVLM & 61.3& 83.0& 65.2& 70.6& 70.1\\
    PyramidDrop & 61.7 & 83.7 & 65.1 & 71.0 & 70.5 \\
    VisionZip & 62.0 & 83.8 & 66.0 & 71.7 & 70.9 \\
    LightVLM & \textbf{64.0} & \textbf{86.4} & \textbf{68.0} & \textbf{73.4} & \textbf{73.0} \\
    \hline
    \end{tabular}
\end{table}

\subsection{Making Larger VLMs Faster Again}
\begin{figure}[t]
    \centering
    \includegraphics[width=\linewidth]{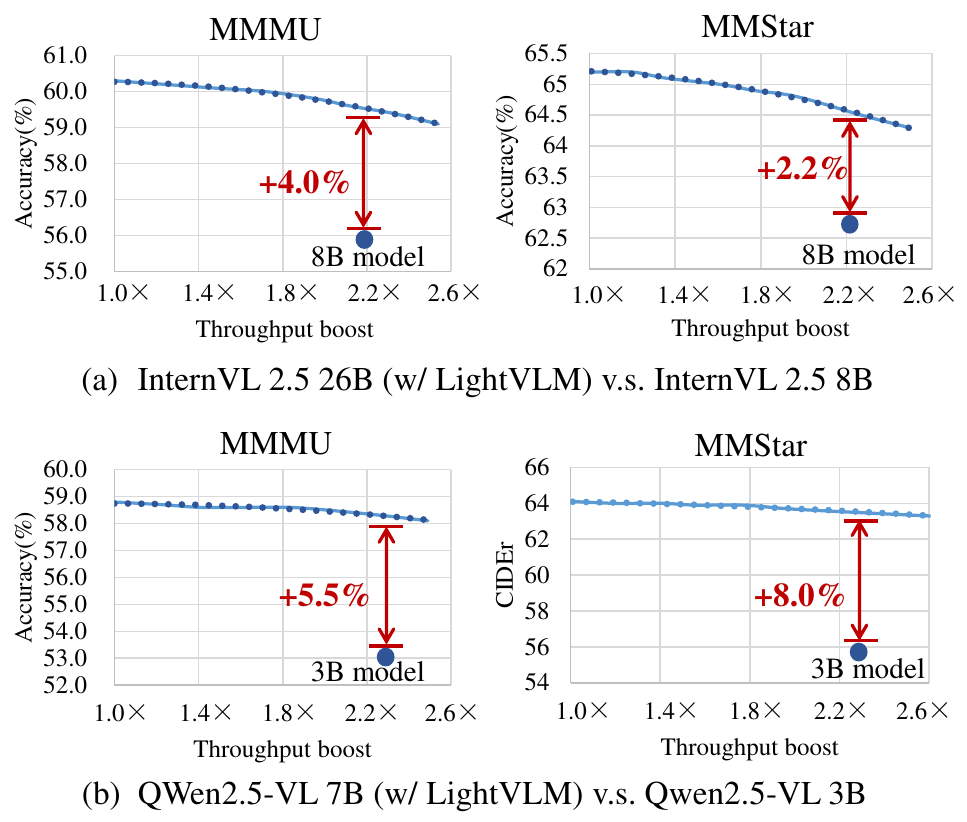} 
    \caption{(a) InternVL2.5 26B w/ LightVLM performs better and runs faster than InternVL2.5 8B. (b) QWen2.5-VL 7B w/ LightVLM performs better and runs faster than QWen2.5-VL 3B.}
    \vspace{-5px}
    \label{fig5}
    \vspace{-5px}
\end{figure}

An impressive advantage of LightVLM is that it could enable a larger VLM to perform better and run faster than smaller VLMs. In Fig.~\ref{fig5}, we deploy LightVLM upon InternVL2.5 26B~\cite{chen2024expanding} and compare it with InternVL2.5 8B~\cite{chen2024expanding}, and deploy LightVLM upon QWen2.5-VL 7B~\cite{bai2025qwen2} and compared it with QWen2.5-VL 3B~\cite{bai2025qwen2}. We notice that with LightVLM, InternVL2.5 26B could own higher throughput than InternVL2.5 8B, while offering 4.0\% and 2.2\% performance advantages over MMMU and MMStar benchmarks. Also, we observe that with LightVLM, QWen2.5-VL 7B achieves higher throughput than QWen2.5-VL 3B, and brings 5.5\% and 8.0\% performance advantages over MMMU and MMStar benchmarks. We conclude that LightVLM provides a new choice to give more reliable and quick responses in real-world scenarios, which can make larger VLMs perform both better and run faster than smaller VLMs.

\subsection{Efficiency Analysis of LightVLM}
We conduct a comprehensive efficiency analysis over network throughput and prefilling time by comparing LightVLM with other efficient methods upon MMMU~\cite{yue2024mmmu} benchmarks in Tab.~\ref{tab6}. We conduct experiments on a single NVIDIA A800-80GB. “Prefilling time” refers to the latency required to generate the first token. Results show that our method not only outperforms other efficient methods over accuracy, but also owns a clear advantage by driving VLMs faster when retaining the same number of image tokens. Notably, LightVLM could increase the network throughput by 2.02$\times$ and decrease the prefilling times by 3.65$\times$, largely increasing the network throughput.
\begin{table}[t]
    \small	
    \setlength\tabcolsep{3pt}
    \centering
    \caption{Efficiency analysis of LightVLM. }
    \vspace{-5px}
    \label{tab6}
    \begin{tabular}{c|ccccc}
    \hline 
    Method & \makecell[c]{Throughput\\(images/s)} & $\Delta$ & \makecell[c]{Prefilling\\ time} & $\Delta$ \\
    \hline
    \rowcolor{gray!20}
    QWen2.5-VL 7B & 1.86 & - & 428ms & - \\
    \hline
    PyramidDrop & 2.97 & 1.60$\times$ & 237ms &  1.80$\times$\\
    SparseVLM & 3.07 & 1.65$\times$ & 251ms & 1.71$\times$\\
    FasterVLM & 3.31 & 1.78$\times$ & 223ms & 1.92$\times$ \\
    VisionZip & 3.83 & \textbf{2.06$\times$} & 96ms & \textbf{4.46$\times$}\\
    LightVLM & 3.75 & 2.02$\times$ & 117ms & 3.65$\times$\\
    \hline
    \end{tabular}
    \vspace{-5px}
\end{table}

\begin{figure}[t]
    \centering
    \includegraphics[width=\linewidth]{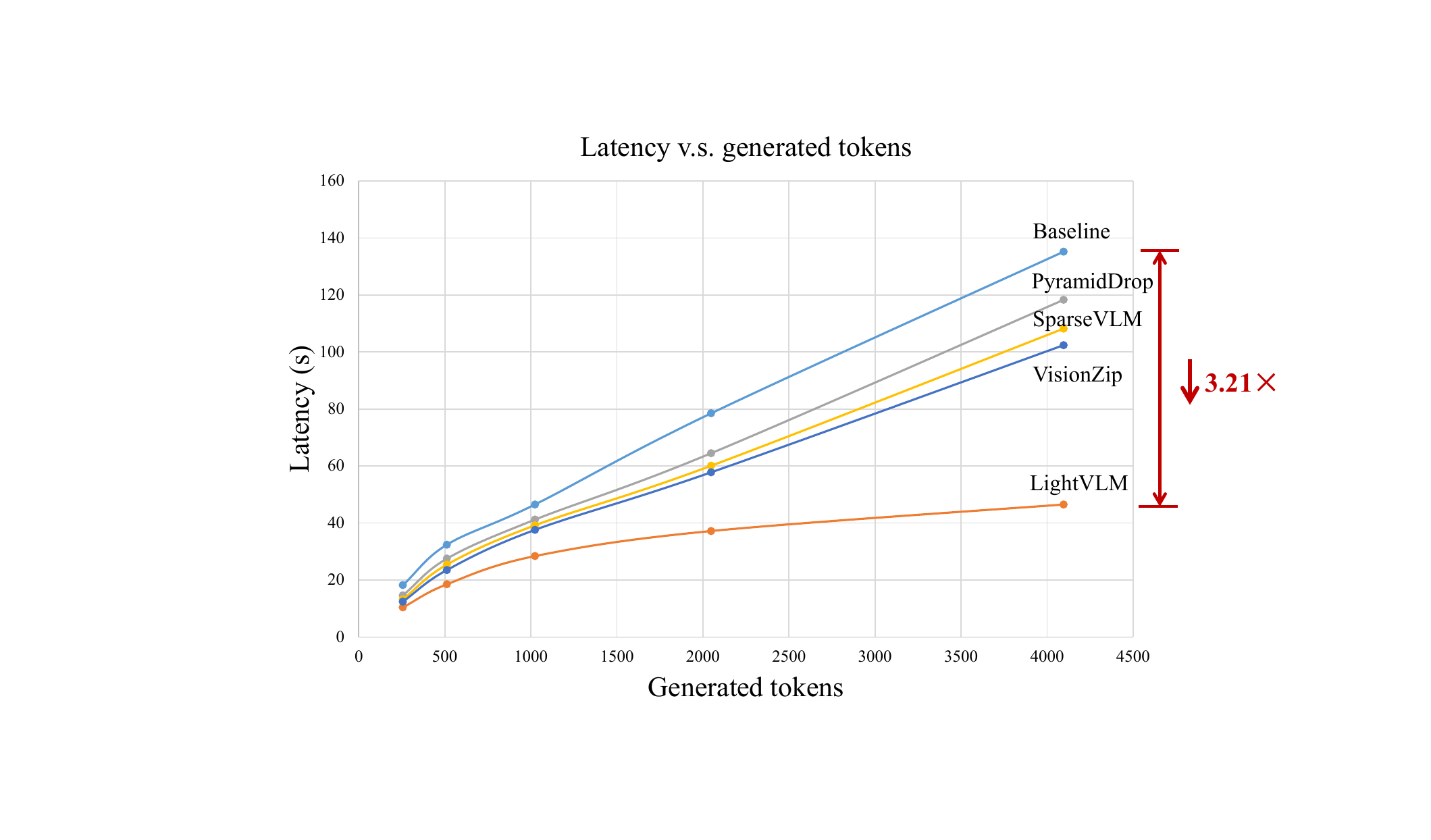} 
    \caption{We measure the latency of generating long output sequences by comparing LightVLM with other efficient methods.}
    \label{fig7}
    \vspace{-10px}
\end{figure}

\begin{figure}[t]
    \centering
    \includegraphics[width=\linewidth]{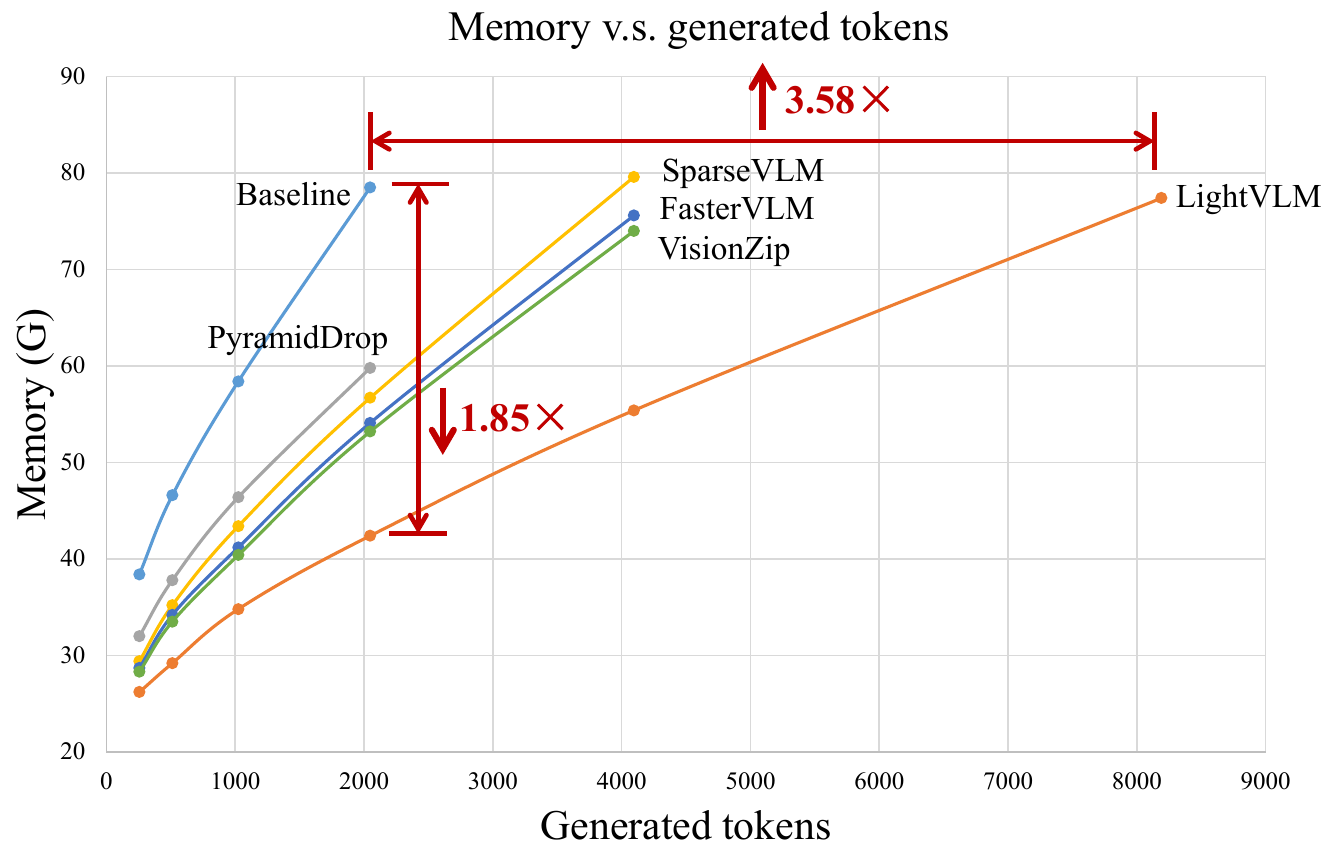} 
    \caption{We measure the memory usage of generating long output sequences by comparing LightVLM with other efficient methods.}
    \label{fig7_sup}
    \vspace{-15px}
\end{figure}

\subsection{Effects When Generating Long Sequences}
We propose KV Cache compression to handle the high latency brought by long output sequences. In Fig.~\ref{fig7}, we measure the latency when generating output sequences of different lengths by comparing LightVLM with other efficient methods. We observe that compared to the vanilla model, existing methods could considerably reduce the generation time but still face high latency. When outputting short sequences (e.g., 128 and 256 tokens), LightVLM notably decreases the latency compared to other methods. When decoding much longer sequences (e.g., 2048 and 4096 tokens), LightVLM demonstrates much clearer advantages compared to other methods. When generating 4096 output tokens, LightVLM notably reduces the network latency by 3.21$\times$ compared to the vanilla model.

In Fig~\ref{fig7_sup}, we measure the memory usage when generating output sequences of different lengths by comparing LightVLM with other efficient methods. We observe that compared to the vanilla model, while existing methods could considerably reduce the memory usage and improve the output length under similar computing budgets, our method is able to demonstrate much superior performance. Compared to the vanilla model, LightVLM decreases the memory consumption by 1.85$\times$ when generating output sequences with the same length (2096). Under the same memory upper limit (80G), LightVLM can significantly boost the generated token sequence length by 3.58$\times$, which verifies the efficacy of our method upon handling long sequence generation.  

\subsection{Visualizations for Preserved Image Tokens}
We provide visualizations for the token merging procedure to show how the kept image tokens preserve various degrees of important information across different layers. Due to the page limit, we show the visualizations in Fig.~\ref{fig8} in the appendix. We observe that VLMs tend to focus more on informative areas such as regions containing texts or meaningful objects. As affordable tokens decrease, VLMs allocate major computations to areas consisting of the most critical details of input images. We conclude that VLMs learn to intelligently distribute computations to aggregate vital information as much as possible to give accurate responses. 

\vspace{-5px}
\section{Conclusion}
In this paper, we conduct a deep analysis of image feature redundancy in Vision-Language Models (VLMs) and utilize it to accelerate modal inference including both encoding and decoding stages. We propose pyramid token merging and KV Cache compression to accelerate VLMs in different stages, respectively. Results across different model sizes over 10+ benchmarks validate the effectiveness and flexibility of LightVLM. LightVLM could 2.02$\times$ the throughput and reduce the latency by 3.21$\times$ when outputting long sequences. 

\section*{Acknowledgments}
This work is supported by National Key Research and Development Program of China (2023YFF0906200) and National Natural Science Foundation of China (Project No. 62476196 and No. 62276182)

\section*{Limitations}

The limitations of this work lie in the following aspects. First, though the token merging algorithm is
compatible with both dot-producted attention and Flash-Attention, we haven't tested it with other efficient attention methods. This is a problem worth further exploration. Second, though we have deployed the proposed LightVLM upon multiple VLMs, there still exist some different VLMs that are not included in our experiments. The scalability of our method could be further verified. 

\bibliography{ref}

\appendix

\section{Appendix}

\begin{figure*}[!h]
    \centering
    \includegraphics[width=\linewidth]{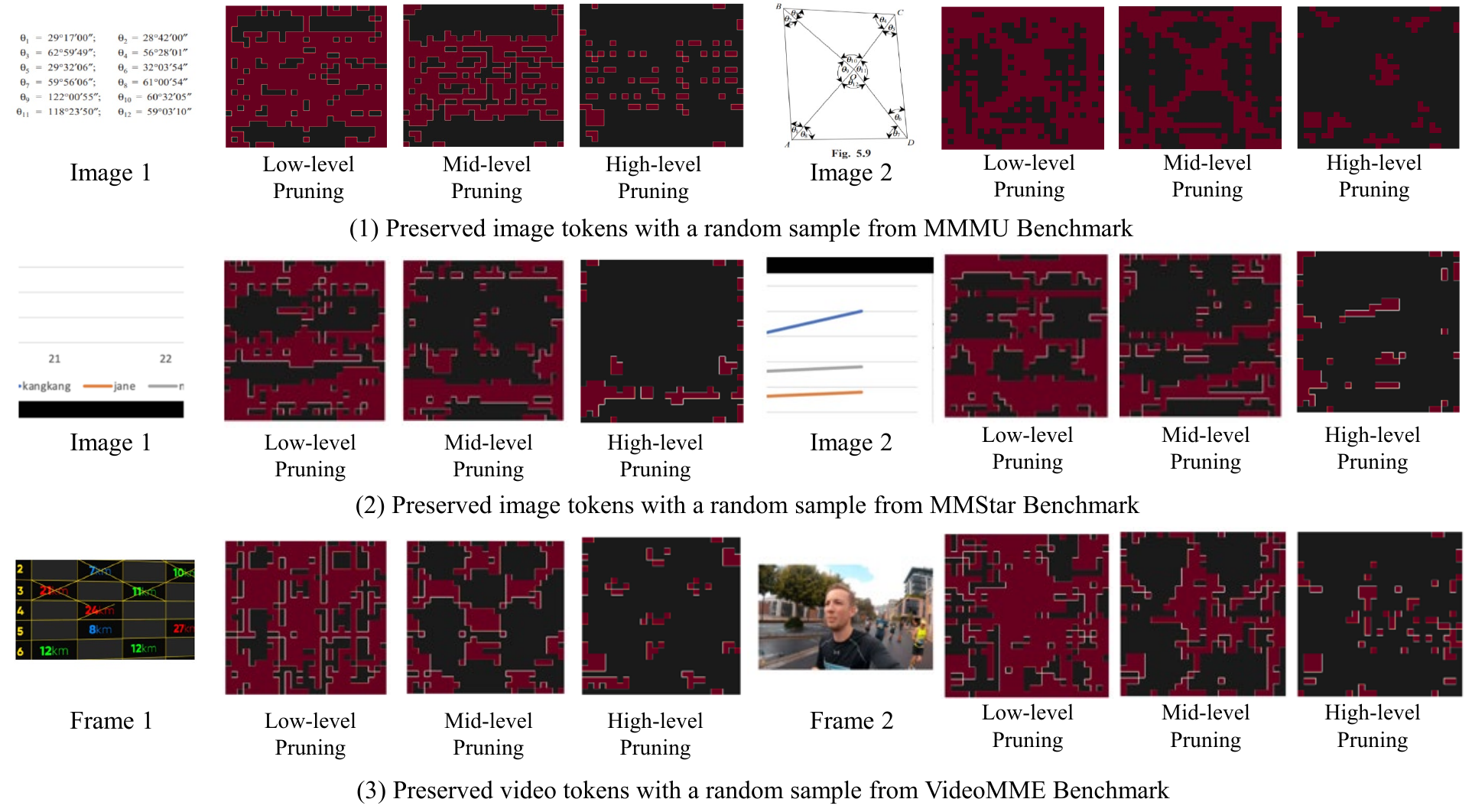} 
    \caption{Visualizations for the preserved image tokens after pyramid token merging compared to original images with a random sample from (a) MMMU benchmark, (b) MMstar benchmark, and (c) VideoMME benchmark. }
    \label{fig8}
    \vspace{-7px}
\end{figure*}

\subsection{Visualizations for Preserved Visual Tokens}
In Fig.~\ref{fig8}, we visualize the preserved image tokens after several LLM layers with pyramid token merging upon QWen2.5-VL 7B~\cite{bai2025qwen2} by three random samples from MMMU~\cite{yue2024mmmu}, MMStar~\cite{chen2024we} and VideoMME~\cite{fu2024video} benchmarks. In Fig.~\ref{fig8}(a), we observe that VLMs tend to focus more on areas such as formulas, angles and lines in images. As affordable tokens decrease, VLMs allocate major computations to areas including the symbols and numbers in formulas. In Fig.~\ref{fig8}(b), we notice that VLMs pay major attention to the legends, digits and curves in images in the initial LLM layers, and gradually focus on meaningful texts or curves when allocatable tokens decrease. In Fig.~\ref{fig8}(c), we find that VLMs could mostly focus on the texts, meaningful objects or the people in the frames which contain the most important information in the scenes. We conclude that VLMs learn to intelligently distribute computations to aggregate vital information as much as possible to give accurate responses. 

\subsection{Model Settings}
By default, we use QWen2.5-VL 7B~\cite{bai2025qwen2} as the backbone model. We perform token merging in Layer 5, 9 and 13 with a constant token reduction ratio which is determined by the computing budget. KV Cache pruning is performed after Layer 5 with the attention threshold $\beta$=0.995.

\subsection{Effects of each proposed component}
We validate the effectiveness of each proposed component by generating long sequences from 512 to 4k. As demonstrated in the following results, each proposed strategy (visual token reduction or KV Cache compression) could notably reduce the inference latency. Meanwhile, we notice that KV Cache compression could decrease more inference latency compared to visual token reduction when output sequences become longer.

\begin{table}[t]   
    \centering
    \setlength\tabcolsep{2pt}
    \caption{The latency (ms) by equipping different proposed strategies.} 
  \label{tab_openasl}
  \begin{tabular}{l|ccccc}
  \hline 
  Vanilla & 32.4 & 46.5 & 78.5 & 135.2\\
  \hline
    w/ token merging & 22.4 & 34.5 & 51.3 & 69.8\\
    w/ KV Cache compresion & 25.2 & 35.4 & 48.6 & 62.8 \\
    LightVLM & \textbf{18.5} & \textbf{28.4} & \textbf{37.2} & \textbf{46.5}\\
  \hline
  \end{tabular}
\end{table} 

\end{document}